\documentclass[letterpaper,twocolumn]{article}
\usepackage{aaai16}
\usepackage{times}
\usepackage{helvet}
\usepackage{courier}

\nocopyright

\usepackage{graphicx}
\usepackage{enumitem}
\usepackage{amsmath}
\usepackage{color}

\usepackage{bigstrut}
\usepackage{array}

\usepackage{natbib}

\renewcommand\cite{\citep}

\newcommand{\eat}[1]{}

\mathchardef\mhyphen="2D
\newenvironment{ite}{                     
     \parskip 0cm \begin{itemize} \parskip 0cm \parsep 0cm \itemsep 0cm \topsep 0cm}{
        \end{itemize}} 

\usepackage{quoting}

\title{Think you have Solved Direct-Answer Question Answering?\\
Try ARC-DA, the Direct-Answer AI2 Reasoning Challenge}

\author{
  Sumithra Bhakthavatsalam,
  Daniel Khashabi,
  Tushar Khot,
  Bhavana Dalvi Mishra, \vspace{1mm} \\
  {\bf \Large
  Kyle Richardson,
  Ashish Sabharwal,
  Carissa Schoenick,
  Oyvind Tafjord,
  Peter Clark} 
         \\
         \ \\
Allen Institute for Artificial Intelligence, Seattle, WA, U.S.A.\\
  {\tt \{sumithrab,danielk,tushark,bhavanad,kyler,ashishs,carissas,oyvindt,peterc\}@allenai.org}
  }

\date{\today}

\begin{document}
\maketitle

\begin{abstract}
We present the ARC-DA dataset, a direct-answer (``open response'', ``freeform'') version of the ARC (AI2 Reasoning Challenge) multiple-choice dataset.
While ARC has been influential in the community, its multiple-choice format is unrepresentative of
real-world questions, and multiple choice formats can be particularly susceptible to artifacts.
The ARC-DA dataset addresses these concerns by converting questions to direct-answer format
using a combination of crowdsourcing and expert review.
The resulting dataset contains 2985 questions with a total of 8436 valid answers (questions typically have more than one valid answer).
ARC-DA is one of the first DA datasets of natural questions that often require reasoning, and where appropriate question decompositions
are not evident from the questions themselves. 
We describe the conversion approach taken, appropriate evaluation metrics, and several strong models.
Although high, the best scores (81\% GENIE, 61.4\% F1, 63.2\% ROUGE-L) still leave considerable room for
improvement.
In addition, the dataset provides a natural setting for new research on explanation, as many
questions require reasoning to construct answers.
We hope the dataset spurs further advances in complex question-answering by the community.\footnote{
ARC-DA is available at https://allenai.org/data/arc-da}
\end{abstract}

\begin{figure}[t]
\centerline{
 \fbox{%
   \parbox{1.0\columnwidth}{
     {\small
{\bf MC:} Many animals depend on plants for (A) shelter {\it [correct]} (B) pollination (C) seed dispersal (D) sunlight \\
{\bf DA:} Many animals depend on plants for what? food $|$ shelter 
\vspace{2mm} \\
{\bf MC:} A solution with a pH of 2 can be increased to a pH above 7 by adding (A) an acid. (B) water. (C) a base. {\it [correct]} (D) hydrogen. \\
{\bf DA:} A solution with a pH of 2 can be increased to a pH above 7 by adding what? a base 
\vspace{2mm} \\
What best describes skin? (A) stiff (B) flexible {\it [correct]} (C) brittle (D) hard \\
{\bf DA:} [Rejected: Too ambiguous as a DA question] 
\vspace{2mm} \\
{\bf MC:} Water freezing is an example of a (A) liquid changing to a solid {\it [correct]} (B) solid changing to a liquid (C) gas changing to a solid (D) gas changing to a liquid \\
{\bf DA:} Water freezing is an example of what? liquid changing to a solid $|$ phase transition $|$ change of state of matter $|$ a change in state $|$ state change 
\vspace{2mm} \\
{\bf MC:} How are the stem of a tree and the stem of a flower most similar? (A) Both are soft. (B) Both have thorns. (C) Both support the plant. {\it [correct]} (D) Both have woody bark. \\
{\bf DA:} How are the stem of a tree and the stem of a flower most similar?
both support the plant $|$ support leaves $|$ both carry water  $|$ both carry nutrients $|$ they support the plant 
}}}
}   
\caption{Multiple-choice (MC) questions from ARC, and their direct answer (DA) equivalents in
the new ARC-DA dataset. Alternative DA answers are separated by a $|$. \label{examples}}
\end{figure}

\section{Introduction}

Multiple-choice (MC) datasets are popular and common in the NLP community,
e.g., CommonsenseQA \cite{Talmor2019CommonsenseQAAQ},
OpenbookQA \cite{Mihaylov2018CanAS}, and VCR \cite{Zellers2019FromRT}, in particular
because of the ease of automatic evaluation. However, they have two notable drawbacks:
First, they are unnatural (real-world questions rarely come with answer options).
Second, the multiple-choice format is particularly susceptible to artifacts, where systems
learn short-cuts to obtain a high score \cite{Gururangan2018AnnotationAI}.

Similarly, while there are many NLP datasets of direct-answer questions (also called ``open response'' or ``freeform'' questions),
e.g., SQuaD \cite{Rajpurkar2016SQuAD10}, TriviaQA \cite{Joshi2017TriviaQAAL}, and NaturalQuestions \cite{Kwiatkowski2019NaturalQA},
the majority of these are span-retrieval (``lookup'') 
tasks where a question is matched against a given/retrieved sentence or paragraph to identify an answer span.
The few DA datasets that do target reasoning, e.g., HotpotQA \cite{yang2018hotpotqa}, DROP \cite{Dua2019DROPAR},
and ROPES \cite{Lin2019ReasoningOP}, are crowdsourced, and thus tend to explore a single, specific style of
reasoning in a controlled setting.

What is missing, still, are direct-answer (DA) datasets of natural questions exploring a {\it wide variety} of
problem types and reasoning styles, and where answers are not constrained to be spans of a source text.
This work alleviates this gap by
supplying such a dataset, namely ARC-DA, a direct-answer version of the ARC (AI2 Reasoning Challenge)
multiple-choice dataset \cite{Clark2018ThinkYH}.
Note that ARC-DA questions are not necessarily more difficult than the original ARC questions (we find
scores on ARC-DA are roughly similar to those on ARC), rather they are
more natural, avoiding the multiple-choice format.



The original ARC dataset contained 
questions 
collected from a large number of science exam and quiz sources.
It has proven useful for the community, stimulating new research in
reasoning-based QA, e.g., \cite{Musa2019AnsweringSE,Boratko2018ASC,Ni2019LearningTA,Xie2020WorldTreeVA},
and as of January 2021 has 35 entries
on its leaderboard\footnote{https://leaderboard.allenai.org/arc/submissions/public}.
ARC is particularly interesting from an NLP perspective: 
the questions were authored by human experts (e.g., examination boards),
they are sensible and high quality, they avoid the repetition common to
crowdsourced datasets, they are highly varied in both the language
they use and the reasoning skills they are designed to probe, and they
are practical, understandable, and motivating. Arguably, the combination of these
factors makes the dataset a useful ``Grand Challenge'' for the field~\cite{Clark2016MyCI}
(The current top score on ARC-Challenge is 81.1\%, thus still with 
room for improvement). The work here, ARC-DA, thus builds on this, providing
a direct-answer version of part of the ARC dataset. Several examples of original
ARC questions and the ARC-DA versions are shown in Figure~\ref{examples}.


We first describe the method used for the conversion, and then
present baseline scores using strong T5-based models. Evaluating DA questions
poses an additional challenge, compared with scoring MC questions. To address
this challenge, we use both human judgements (obtained with GENIE, an automated crowdscoring pipeline \cite{genie}),
and automated metrics. 
Although high, the best scores (81\% GENIE, 61.4\% F1, 63.2\% ROUGE-L) still leave considerable room for
improvement.
In addition, the dataset provides a natural setting for new research on explanation, as many
questions require reasoning to construct answers.
We encourage the community to make use of this dataset to make further progress in advanced question-answering.

\section{ARC-DA Dataset}

Na\"{i}vely, one can convert MC to DA simply by removing the answer choices,
and using the correct answer choice as the target answer.\footnote{Indeed, this is
  the approach taken by \cite{Lin2020DifferentiableOC} to use (a filtered subset of) ARC in a direct-answer setting.}
However, there are several problems that can arise:
\begin{ite}
\item There may be multiple ways of wording the correct answer.
\item There may be multiple possible correct answers, and in some cases too many to enumerate all of them.
\item The question itself may be ill-defined without answer options.
\end{ite}
To address these problems, we convert the 7787 ARC MC questions to DA using the process described below.

\subsection{Crowdworker Annotation}

We start with a large scale crowdsourcing process to filter questions to those suitable for the DA setting and collect alternative correct answers for them:

\noindent
\begin{enumerate}
\item      {\bf Initial Question Filtering:} Remove questions where the question sentence\footnote{
  Many questions are multi-sentence, with a preamble before the actual question sentence.}
  contains one of several empirically-chosen filter phrases, e.g., ``Which of''.\footnote{
    The filter phrases are: which of,  most,  best,  least, est,  order,  supports, characteristic,  trait, which object, which statement, below, which is, which are, example, which term,  conclusion, which would, which item, which action, which two, which sentence, which one,  sequence, which fact, which $<${\it VERB}$>$.}
  Questions containing these phrases were observed to usually be ill-formed without the answer options, e.g., ``Which of these items contains only a liquid?''.
\item {\bf Collecting Answers:} Each question was then posed to five independent crowdworkers as a DA question, and the workers were asked to:
 \begin{ite}
 \item Answer the question (enter a free-form answer). If there were multiple answers, they were asked to enter two or three.
 \item Identify if the question had one, several, or many answers, or if the question was nonsensical. 
 \end{ite}
 If the question was too ambiguous or nonsensical, the crowdworker had the option of not providing an answer.
 The crowdworker interface is shown in Appendix~A.
\item {\bf Additional Filtering:} The questions were further filtered, only retaining:
  \begin{ite}
  \item questions that had answers from at least two workers.
  \item questions where at least two worker-provided answers had {\it some} non-stop-word overlap.
  \end{ite}
  Otherwise the question was deemed too open-ended and rejected. 
\end{enumerate}

\subsection{In-House Review}

\noindent
The resulting questions were then reviewed by in-house (``expert'') workers, who performed the following operations:
\begin{enumerate}
 \item {\bf Question Filtering:} Rejected questions that still appeared too open-ended (e.g., ``Name an insect.'').
 \item {\bf Answer Verification:} Reviewed crowdworker answers to remove incorrect answers, and add additional missed answers.
 \item {\bf Question Rewording:} Reworded questions that were poorly phrased or incomplete as standalone questions, e.g.,
       ``The cell structure that makes a plant cell more rigid than an animal cell is the'' becomes
       ``The cell structure that makes a plant cell more rigid than an animal cell is called what?''
 \item {\bf Answer Modification:} For long (wordy) answers, ensure that a shorter version including just the salient terms is
       also present. For example, for the question: ``In what form does water vapor exist in the atmosphere?'',
       the crowdworkers gave two answers: ``An invisible gas in the air'', and ``An invisible gas''.
       As the simple answer ``gas'' is sufficient for this question, the expert would add ``gas'' as
       an additional answer option.
\end{enumerate}
This process was run over the entire ARC question set. Approximately 60\% of
the original questions were removed during crowdworker annotation (50\% in the initial question filtering,
10\% more in the additional filtering), 
followed by another 10\% during in-house review, resulting in 2985 questions
in the final ARC-DA dataset. Although the final dataset is less that half the size
of ARC, it is still large enough for models to learn the style of the task (e.g., see Table~\ref{results} later),
without simply memorizing the task itself, thus avoiding large-scale supervised training pitfalls.
This trend towards more realistically sized datasets is seen elsewhere also,
e.g., OBQA \cite{Mihaylov2018CanAS}, QASC \cite{khot2019qasc}, TRACIE \cite{Zhou2020TemporalRO}.

\begin{table}[t]
\centering
{\small
\begin{tabular}{lccc} 
                        & {\bf Train} & {\bf Dev} & {\bf Test} \\ \hline
\bigstrut
num.\ questions		& 1250	& 338	& 1397 \\
num.\ answers per qn (avg)     & 2.75	& 2.72	& 2.92 \\
num.\ words per answer (avg)	& 2.11  & 1.94  & 2.27 \\ \hline
\end{tabular}
} 
\caption{Statistics of ARC-DA, with 2985 total questions.\label{table:statistics}}
\end{table}

\subsection{Train/Dev/Test Split}

We retain the same train/dev/test labels for questions as in the original ARC dataset, resulting in approximately
similar proportions as ARC. We also do not separate the original ARC-Easy and ARC-Challenge questions,
but instead merge them into a single dataset. We do this
because the labels ``Easy'' and ``Challenge'' were based on the MC choices.
(Switching from MC to DA can result in a ``Hard'' question becoming conceptually easy, and vice versa).
However, we do retain the original Easy/Challenge labels as metadata in the ARC-DA dataset. The resulting dataset statistics are summarized in Table~\ref{table:statistics}.

\subsection{Knowledge and Reasoning Types}

\begin{figure}[t]
\centering
\includegraphics[width=0.95\columnwidth]{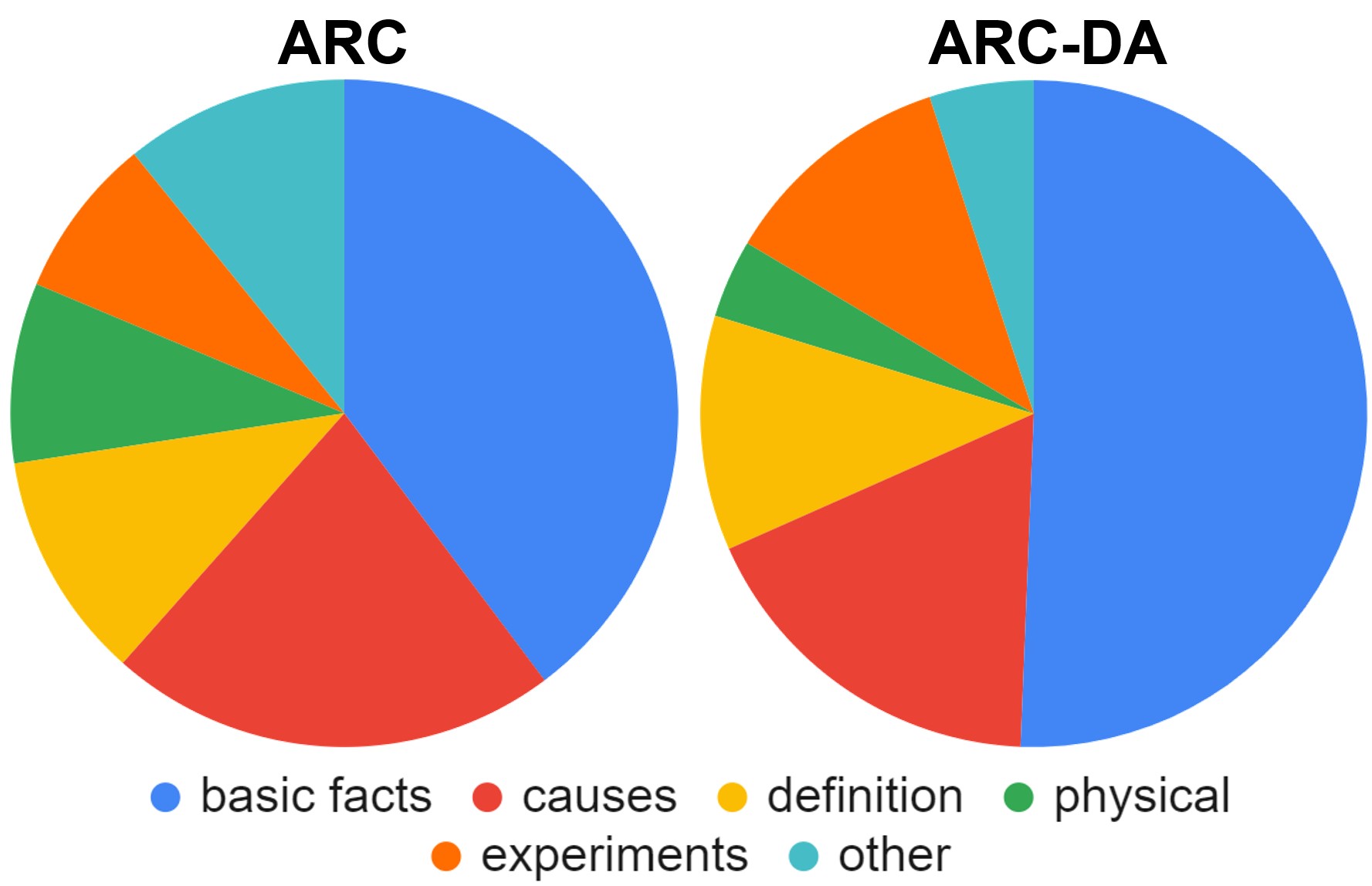} \\
                {\bf Knowledge Types} \\
\ \\
\includegraphics[width=0.95\columnwidth]{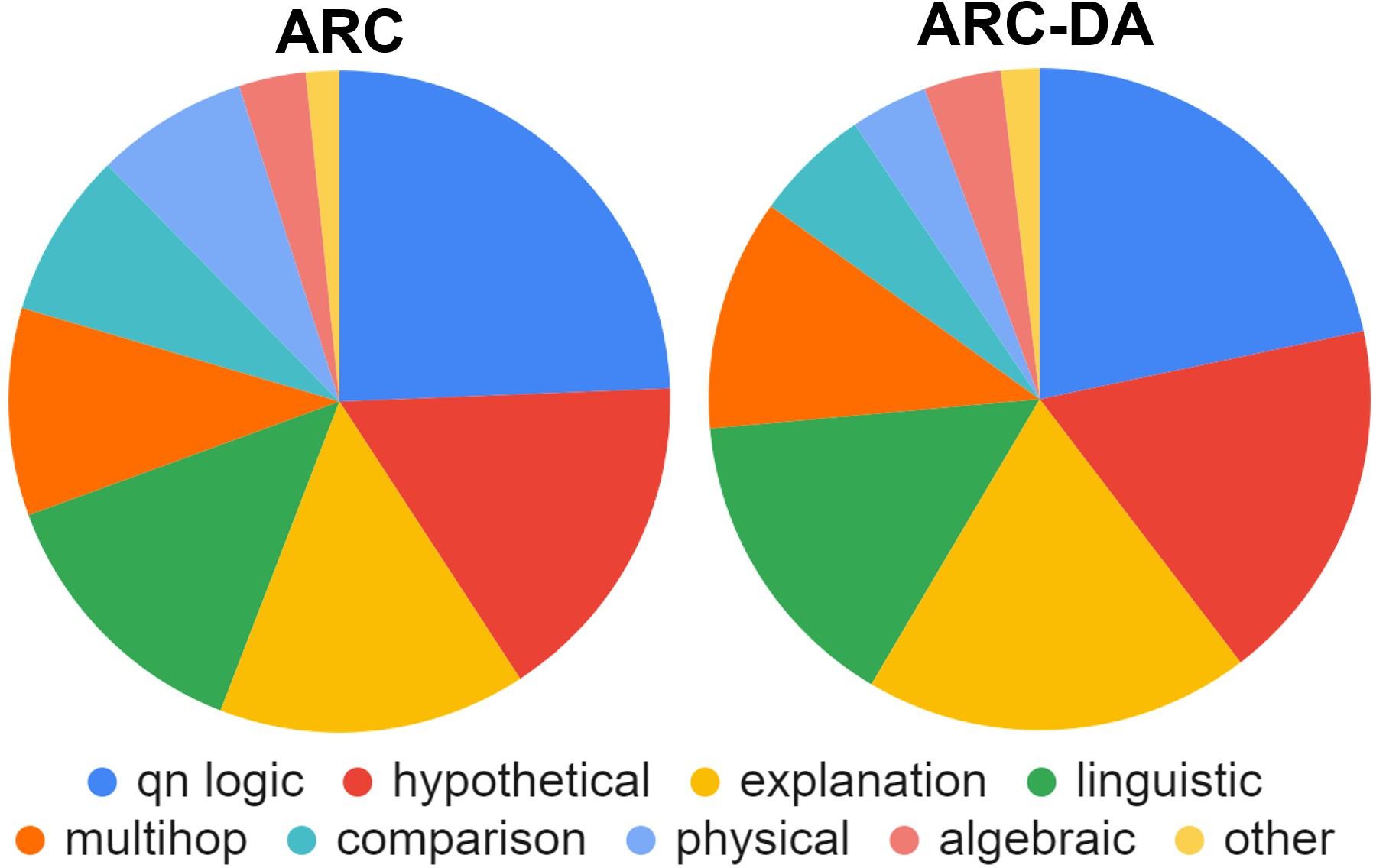} \\
{\bf Reasoning Types}
\caption{Comparison of the distribution of questions among different knowledge (top) and
  reasoning types (bottom), comparing ARC with ARC-DA. Overall, the distributions are
  roughly similar. Data is from sampled annotations created by \cite{Boratko2018ASC}.
  For a detailed description of the categories, see \cite{Boratko2018ASC}.
\label{knowledge-and-reasoning-types}}
\end{figure}

 We found that the distribution of knowledge and reasoning types required by ARC-DA questions, as classified by \citet{Boratko2018ASC}, to
be roughly the same as in ARC, see Figure~\ref{knowledge-and-reasoning-types} (created using Boratko et al's data).
For a detailed description of these categories, see \cite{Boratko2018ASC}.

\begin{table}
  \centering
{\small
\begin{tabular}{ll}
  \hline
      {\bf Rating} & {\bf Score} \\ \hline
      strongly agree & 1.00 \\
      agree & 0.75 \\
      neutral & 0.50 \\
      disagree & 0.25 \\
      strongly disagree & 0.00 \\ \hline
\end{tabular}
}
\caption{GENIE's crowdworker ratings of a model's answers are mapped to real-value scores as shown. \label{genie-rubric}}
\end{table}

\subsection{Evaluation Metrics}

It's not immediately clear how one should score answers to DA questions. Doing this is more difficult than for MC questions, as (usually) the set of gold DA answers is incomplete. Further, even if the answer is unique conceptually (e.g., the answer ``gravity'') it may be phrased in multiple ways (``the force of gravity'' ``gravitational force'', ``gravitation'', ...). As a result, scoring is necessarily approximate. However, this should not be a reason to shy away from such problems; valid comparisons can still be made, and there are obvious benefits to working in the more realistic DA setting.

We propose two ways to score answers to ARC-DA: The first is human scoring via GENIE\footnote{Available at https://genie.apps.allenai.org/}, a human-in-the-loop leaderboard framework that scores answers using an automated crowdsourced pipeline \cite{genie}. GENIE streamlines the human scoring of machine-generated answers by automatically posting them on crowdsourcing platforms, collecting qualitative human judgements (converted to numeric scores using the rubric in Table~\ref{genie-rubric}),
then performing statistical analyses to quantify uncertainty. It also includes various constraints to ensure quality control. To use GENIE, we submit our answers to the leaderboard, then wait for the task to complete (which follows a fixed, periodic schedule). Note that GENIE is publicly available for other researchers interested in this dataset.

Second, we consider two popular automatic metrics to score answers by comparing them to the (typically incomplete) set of gold answers, namely ROUGE and an F1 word-overlap measure.

For ROUGE \cite{Lin2006AnIA}, we use the F1 score for the ROUGE-L variant which considers the longest common subsequence, thus penalizing words out of order.\footnote{We use the implementation from https://github.com/google-research/google-research/tree/master/rouge, with stemming turned on.} For the simple F1 word-overlap measure, we adopt the conventions from the SQuAD dataset \cite{Rajpurkar2016SQuAD10} in terms of ignoring punctuation and a few stop words. For both ROUGE and F1, we take the maximum score over all of the gold answers for a given question (i.e., an answer is scored against its best-matching gold answer), and then average over all the questions.

We note that both ROUGE and F1 have known intrinsic pitfalls. For example, as F1 ignores word order, the prediction ``from solid to liquid'' would be considered a perfect match for the gold answer ``from liquid to solid''.

For these reasons, our preferred metric for ARC-DA is GENIE (despite the turnaround time), which also alleviates the problem of missing gold answers.

\section{Empirical Evaluation}

We next describe a few strong baseline systems for ARC-DA and report their performance.

\subsection{Baseline Models}

To build a strong baseline model, we start with (a reimplementation of) UnifiedQA \cite{Khashabi2020UnifiedQACF}, a QA system
trained on multiple QA datasets using the text-to-text pretrained T5 transformer \cite{Raffel2020ExploringTL}
(we use the 11B version). We then fine-tune two models on ARC-DA, one using sentences retrieved from a general corpus
of text $K$, and one without. The input to these models is the question $Q$ (plus retrieved sentences, for the first
model). The desired output is a correct answer to $Q$. We call the resulting models {\bf UnifiedQA + ARC-DA}.

For the {\bf ``with IR''} (Information Retrieval) variant of UnifiedQA + ARC-DA, given a question $Q$, 
we retrieve 10 sentences $K_{1},...,K_{10}$ from the corpus $K$ using $Q$ as the search query (here, using ElasticSearch).
For $K$, we use the Aristo Corpus, a Web-crawled corpus containing 280GB of general and
science-related sentences augmented with $\approx$80k additional science textbook sentences \cite{Clark2016CombiningRS}.
The input to the model is then:
  \begin{quote}
    {\tt \$question\$ = } {\it Q} {\tt ; \$context\$ = } $K_{1} ... K_{10}$ 
  \end{quote}
  The desired output of the model is a correct answer to the question.
  To train the model, since we (typically) have multiple, alternative gold target answers $A_{1},...,A_{n}$ in the training data, we generate $N_a$ training examples for each question, where each example uses a randomly sampled answer from $A_{i}$. In other words, each individual gold answer (of which there are a few per question) and unique question are used to construct an individual training example, capped at a max of $N_{a}$ training examples per question. In our experiments, we used $N_a = 4$. 
  \eat{
  For the target answer (output to be generated), as there are multiple, alternative answers $A_{i}$ for each question, we create 4 training
  examples for each question, each with the same input but with a different target answer $A_{i}$. To do this, we sample the $A_{i}$ four times, each
  time picking a different random answer (unless there are less than 4, in which case answers will be repeated).
  Thus if there are more than 4 answers, we only sample 4 of them for the training data.
  }
Each training instance thus has a single gold answer, and the fine-tuning otherwise follows the T5 procedure of using teacher forcing \cite{Williams1989ALA}. Note there is a (deliberate) asymmetry in train/test: Each training instance encourages the system to predict a {\it particular} gold answer, while each test output is considered correct if it predicts {\it any} of the gold answers. This style of teaching for questions
with multiple answers has been found effective in previous work, e.g., 
\cite{Bosselut2019COMETCT,Rashkin2018ModelingNP}.

For the {\bf ``without IR''} variant, the same process is applied except the input to the model is simply:
  \begin{quote}
    {\tt \$question\$ = } $Q$
  \end{quote}
  
  Since UnifiedQA is question-format agnostic,\footnote{
That is, given an MC question, UnifiedQA will output an answer choice label; while given a DA question, UnifiedQA will generate an answer directly.}
    we also create variants of the above models (again with and without retrieval) by fine-tuning them jointly on ARC-DA as described above as well as on the original multiple choice questions of ARC. The resulting models are referred to as {\bf UnifiedQA + ARC-DA/MC}.

    \begin{table}[t]
\centering
{\small
\setlength{\tabcolsep}{3pt}	
\begin{tabular}{lccc}
 & \multicolumn{3}{c}{\bf Score (Test Set)} \\
{\bf Model:}	& {\bf GENIE}	& {\bf F1}  & {\bf ROUGE-L}	\\ \hline
T5 + ARC-DA (no IR)        & 66$^{+3}_{-3}$    &           & 50.0 \\
UnifiedQA + ARC-DA (no IR)    & 72$^{+2}_{-3}$    & 53.5		& 55.7	\\ 
UnifiedQA + ARC-DA (w/ IR)   & 75$^{+2}_{-2}$    & 59.6		& 61.2	\\ 
UnifiedQA + ARC-DA/MC (no IR)    & 75$^{+2}_{-2}$    & 55.4		& 57.5	\\ 
UnifiedQA + ARC-DA/MC (w/ IR)   & {\bf 81}$^{+2}_{-2}$   & {\bf 61.4}	& {\bf 63.2}	\\ \hline
\end{tabular}
}
\caption{Results on ARC-DA test set (1397 questions), both without and with IR, according to different metrics.
  GENIE is a human (crowdsourced) metric, F1 and ROUGE-L are automated metrics.
  The GENIE score includes a confidence interval (+/-), as shown.
  (GENIE is our preferred measure.) \label{results}}
\end{table}

\begin{table}[t]
\centering
{\small
\setlength{\tabcolsep}{3pt}	
\begin{tabular}{lccc}
 & \multicolumn{3}{c}{\bf Score (Dev Set)} \\
{\bf Model:}	& {\bf EXPERT}	& {\bf F1}  & {\bf ROUGE-L}	\\ \hline
UnifiedQA + ARC-DA (no IR)         	& 78.8   & 53.9		& 55.4	\\ 
UnifiedQA + ARC-DA (w/ IR)      	& 84.0   & 63.0		& 65.2	\\
UnifiedQA + ARC-DA/MC (no IR)     	& 78.7   & 55.5		& 59.5	\\ 
UnifiedQA + ARC-DA/MC (w/ IR)   	& {\bf 85.9} & {\bf 63.7}	& {\bf 66.8}	\\\hline
\end{tabular}
}
\caption{Results on ARC-DA dev set (338 questions). Here we show human evaluation by one of the authors (EXPERT), rather than GENIE scores. \label{results-dev}}
\end{table}

\subsection{Results}


The results for the models are shown in Table~\ref{results}. 
To help interpret the GENIE scores, note that crowdworkers label answers according to the rubric and corresponding real values as 
shown in Table~\ref{genie-rubric}. For comparison, one of the authors manually scored the answers on the development set, using a principle of partial credit for non-ideal answers; this is shown under the EXPERT column of Table~\ref{results-dev}.

There are several results of note.
First, {\bf the scores are high} in absolute terms, with the human-scored GENIE/EXPERT numbers being roughly comparable to scores on the original
MC questions, found to be 86.8\%/92.6\% without/with IR.\footnote{
  To obtain these MC scores, we ran the same UnifiedQA model, before fine-tuning on ARC-DA, on the original ARC multiple-choice versions
  of the 1397 ARC-DA test questions.}
This suggests that the DA questions are not necessarily harder than the MC versions, despite the format change,
although they are more natural (non-multiple-choice).
While intuitively one might expect DA questions to be more difficult to answer as the number of potential answers changes 
from 4 to a potentially infinite number, some may also be easier as {\it any} correct answer is valid,
allowing the model to sidestep subtle distinctions that may be used in the MC choices.


Second, the {\bf GENIE scores slightly underestimate the ``true'' score}, which
we take as the EXPERT score (Table~\ref{results-dev}), namely the score one might
expect to receive in an examination setting with a professional grader. This
may be due to occasional annotation errors and/or unreliable annotators that slip through
GENIE's quality controls. 
(Also note the GENIE score in Table~\ref{results} is on the test set, while the EXPERT score
in Table~\ref{results-dev} is on dev, which may account for some of the difference (test performance
is typically slightly worse than dev)). 
While in principle the upper bound on the EXPERT
score is 100\%, namely for a perfect set of answers, our preliminary tests
suggest the GENIE upper bound (for ARC-DA) may be around 90\% for a perfect set of answers
due to this noise, given GENIE's current pipeline (additional improvements to GENIE are
under consideration).

Third, the {\bf automated metrics are only a loose approximation} of the true target.
In absolute terms, there is a significant gap between
the automated metrics (F1 and ROUGE-L) and the human evaluations (GENIE and EXPERT), suggesting that
there are indeed additional answers and answer phrasings missing in ARC-DA gold answers.
We also see that the rank-ordering of models based on human vs. automated metrics
is not identical (although is generally similar).
Assuming that the human-based scores are the most accurate (although expensive),
this indicates that automatic metrics should be used with caution: While they can be used as
a useful proxy, it is not appropriate to draw conclusions from them based
on small (e.g., 1\%) differences.



\eat{
Note that the rank ordering is the same (although in general this may not be so, depending
on the quality, completeness, and linguistic variability of the gold answers).}

\subsection{Impact on MC Question-Answering}

As an unexpected corollary, we ran the UnifiedQA + ARC-DA/MC model on the original ARC MC dataset,\footnote{
  As before, note that UnifiedQA is format-agnostic, outputing an answer option label given an MC question, or a direct answer given a DA question.}
and obtained new state-of-the-art results (81.4\% on ARC-Challenge and 92.7\% on ARC-Easy).\footnote{https://leaderboard.allenai.org/arc/submissions/public}
Note also that this model has the highest score on ARC-DA (GENIE score of 81\%, Table~\ref{results}).
This suggests that there is some additional training signal provided by the DA training questions that is assisting in MC QA,
and likewise that the additional MC training is helping answer DA questions. This phenomenon is
reminiscent of the discovery in the original UnifiedQA paper that multi-format training can provide an overall boost in individual scores \cite{Khashabi2020UnifiedQACF}.

\eat{
\section{Discussion}

\subsection{DA vs. MC}

A perhaps surprising result is that the absolute GENIE scores are quite high in absolute terms (72\%-75\%),
countering the intuition that DA is significantly harder than MC.
For comparison, the same UnifiedQA model (before fine-tuning on ARC-DA) scores 86.8\% (without IR) and 92.6\% (with IR)
on the original ARC multiple-choice versions of the 1397 ARC-DA test questions.\footnote{
  On {\it all} the ARC test set (3548 questions), these MC scores are slightly lower at 83.1\% and 88.7\% respectively, indicating that the ARC-DA filtering has kept a larger fraction of ``easier'' questions. }
 Although DA questions are in one sense harder than MC (as the number of potential answers changes
from 4 to a potentially infinite number), some may also be easier as {\it any} correct answer is valid,
allowing the model to sidestep subtle distinctions that may be used in the MC choices. 

Note that retrieving relevant text is harder for DA than MC, since the retrieval query in the MC setting can include the answer choices,
typically resulting in more relevant text. This is reflected by the narrower gap between the without/with IR models when comparing the results on DA vs.\ MC.
}


\section{Summary}

Progress in QA requires new datasets in more realistic settings, for
example using natural questions that require more than a ``lookup'' answer. The ARC-DA
dataset addresses this need, containing a direct answer version of (a subset of)
the ARC multiple-choice questions. These questions are expert (examination board) authored,
high quality, sensible, and avoid the repetition common to crowdsourced datasets,
making them of particular interest to NLP. We have also shown that baseline scores,
although strong, are far from perfect, offering a new challenge to the NLP community,
as well as a new setting to study explanation in the context of questions requiring reasoning.
We invite readers to take up this challenge!

The ARC-DA dataset is available at https://allenai.org/data/arc-da, and the
GENIE human evaluation framework is publicly available at https://genie.apps.allenai.org.

\section*{Acknowledgements}

Thanks to all in the Aristo team and the additional expert reviewers Kirsten Barber, Rosann Morrow-Clark, Tao Li, and Anjali Tandon
who contributed to this dataset. The TPU machines for conducting experiments were provided by Google.

\bibliography{references}
\bibliographystyle{myabbrvnat}

\onecolumn


      \begin{figure*}[ht]
  \noindent
      {\large {\bf Appendix A. Instructions to Crowdworkers}}
      \noindent
      Below are the instructions provided to the (Amazon Mechanical Turk) crowdworkers for answering DA questions:

      \vspace{-1cm}
      
      \centering
  \includegraphics[width=1.0\textwidth]{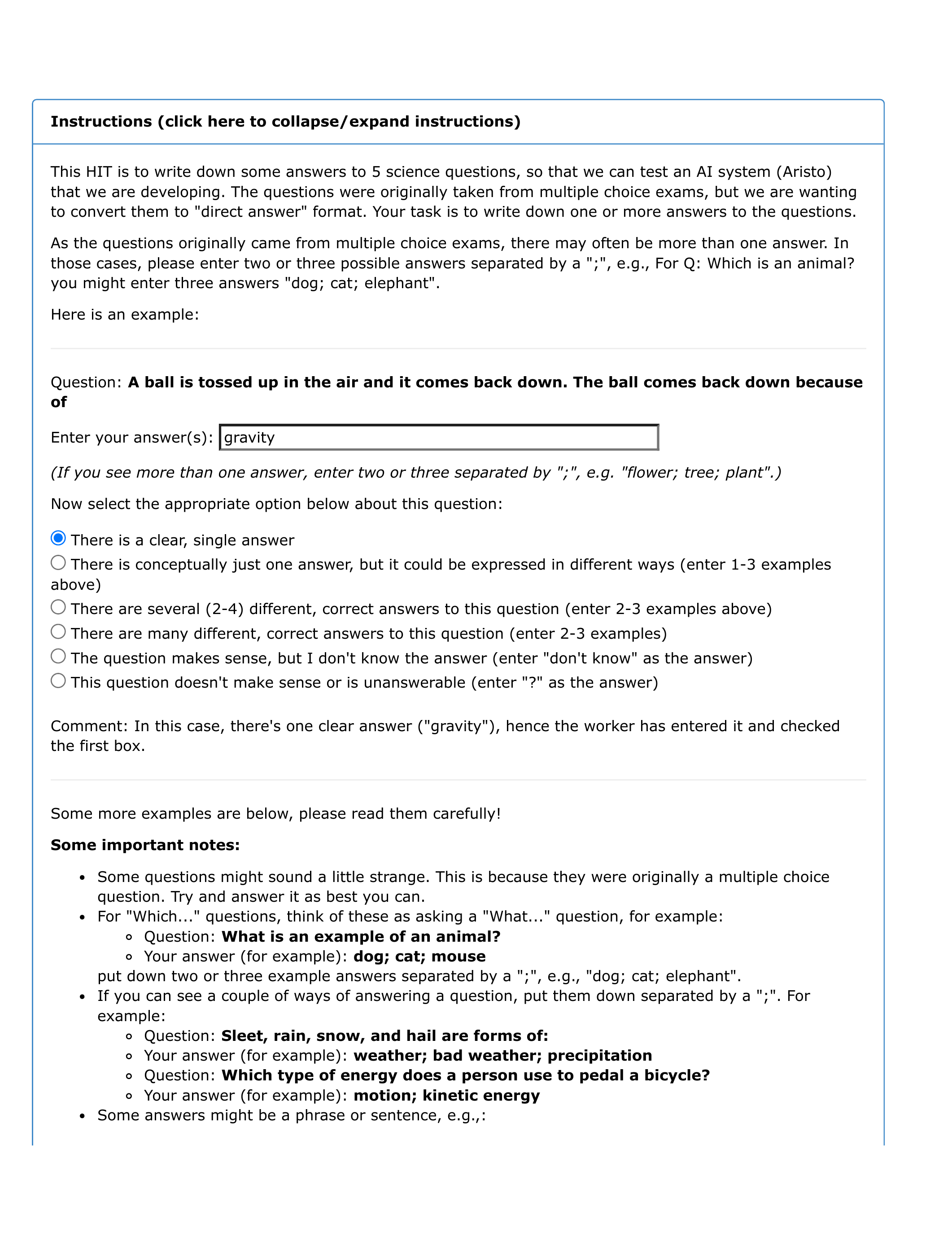}
\end{figure*}

\begin{figure*}[ht]
\centering
  \includegraphics[width=1.0\textwidth]{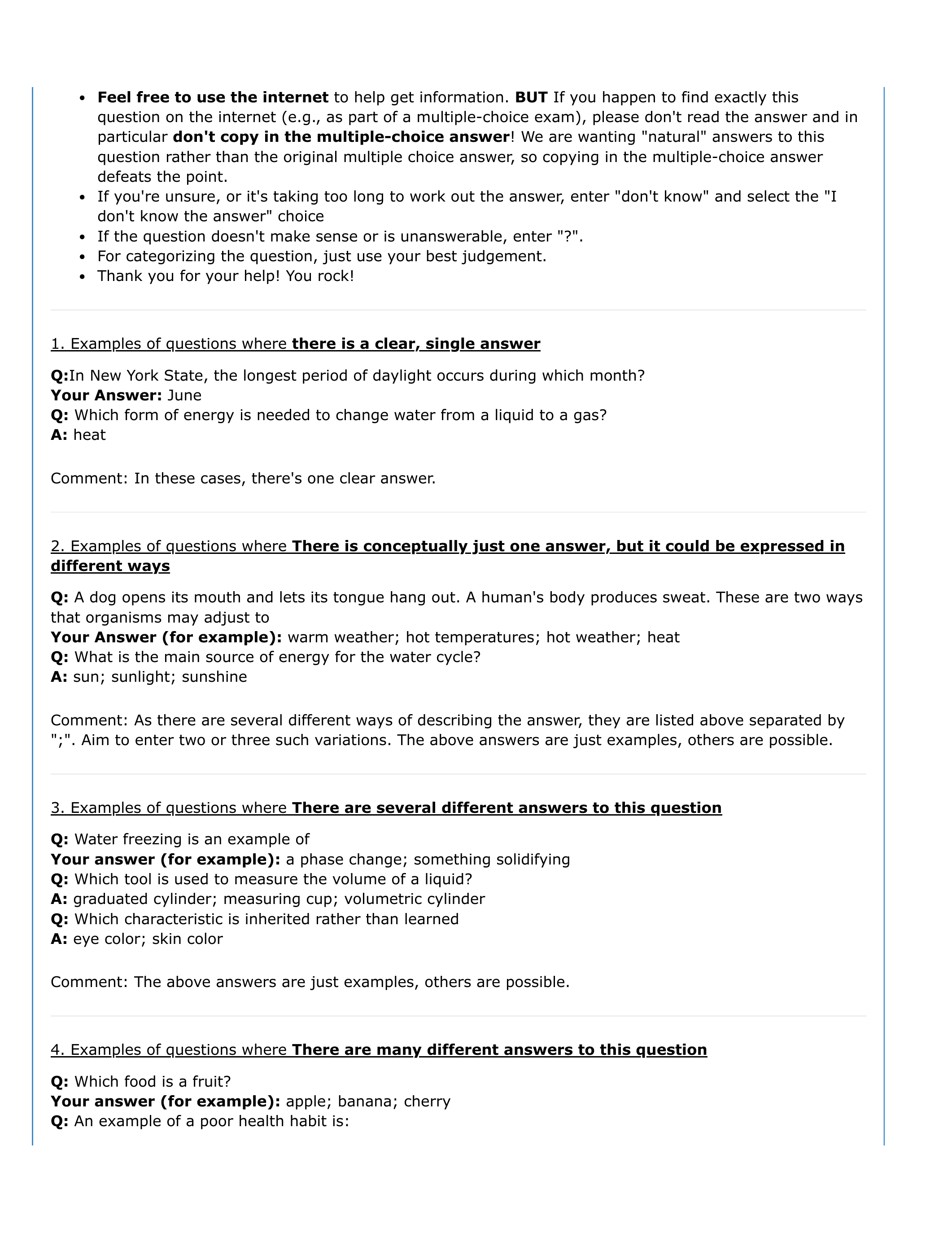}
\end{figure*}

\begin{figure*}[ht]
\centering
  \includegraphics[width=1.0\textwidth]{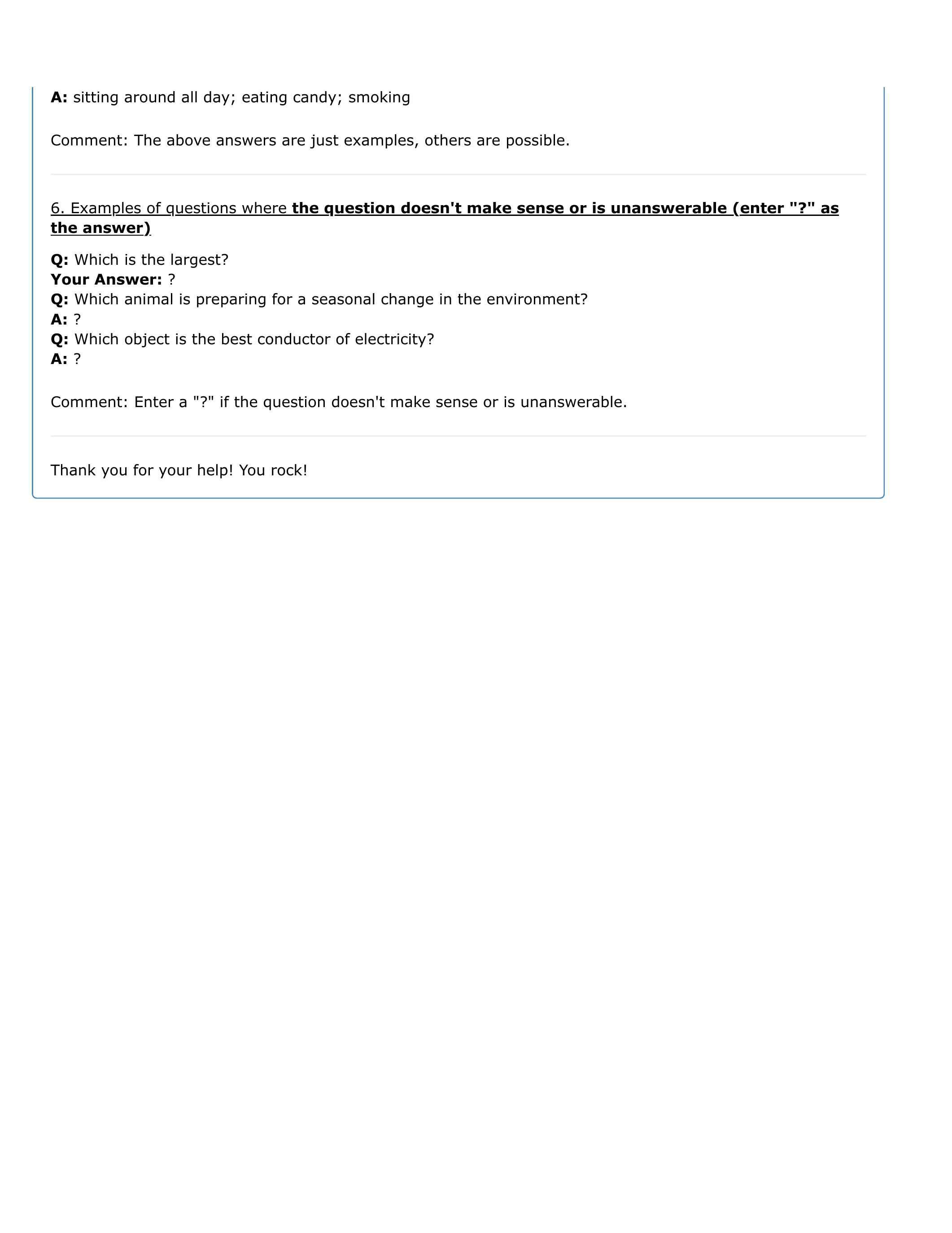}
\end{figure*}

\end{document}